\documentclass{article} 
\usepackage{iclr2026_conference,times}

\usepackage{multirow} 
\usepackage{wrapfig}
\usepackage{amsmath}
\usepackage{graphicx} 
\usepackage[utf8]{inputenc} 
\usepackage[T1]{fontenc}    
\usepackage{hyperref}       
\usepackage{url}            
\usepackage{booktabs}       
\usepackage{amsfonts}       
\usepackage{nicefrac}       
\usepackage{microtype}      
\usepackage{xcolor} 
\usepackage{hyperref}
\usepackage{url}
\usepackage{graphicx} 
\usepackage{algpseudocode}
\usepackage{algorithm}
\usepackage{algpseudocode}
\usepackage{algorithmic}
\usepackage{multirow}
\usepackage{algpseudocodex}
\usepackage[dvipsnames]{xcolor}
\usepackage{amsmath}
\usepackage{amsthm}
\usepackage{tabularx}
\usepackage{booktabs}
\usepackage{array}
\usepackage{enumitem}
\usepackage{subcaption}
\theoremstyle{plain}
\newtheorem{theorem}{Theorem}[section]

\theoremstyle{definition}

\theoremstyle{remark}
\newtheorem{remark}[theorem]{Remark}

\title{Domain-Aware Tensor Network Structure Search}

\author{
Giorgos Iacovides$^{1}$\thanks{Equal Contribution}, Wuyang Zhou$^{1}$\footnotemark[1], Chao Li$^2$, Qibin Zhao$^2$, Danilo Mandic$^1$ \\
$^1$Department of Electrical and Electronic Engineering, Imperial College London \quad $^2$RIKEN AIP \\
\texttt{\{giorgos.iacovides20,wuyang.zhou19,d.mandic\}@imperial.ac.uk} \\
\texttt{\{chao.li, qibin.zhao\}@riken.jp} \\
}

\iclrfinalcopy 
\begin{document}

\maketitle

\begin{abstract}
Tensor networks (TNs) provide efficient representations of high-dimensional data, yet identification of the optimal TN structures, the so called tensor network structure search (TN-SS) problem, remains a challenge. Current state-of-the-art (SOTA) algorithms solve TN-SS as a purely numerical optimization problem and require extensive function evaluations, which is prohibitive for real-world applications. In addition, existing methods ignore the valuable domain information inherent in real-world tensor data and lack transparency in their identified TN structures. To this end, we propose a novel TN-SS framework, termed the tnLLM, which incorporates domain information about the data and harnesses the reasoning capabilities of large language models (LLMs) to \textit{directly} predict suitable TN structures. The proposed framework involves a domain-aware prompting pipeline which instructs the LLM to infer suitable TN structures based on the real-world relationships between tensor modes. In this way, our approach is capable of not only iteratively optimizing the objective function, but also generating domain-aware explanations for the identified structures. Experimental results demonstrate that tnLLM achieves comparable TN-SS objective function values with much fewer function evaluations compared to SOTA algorithms. Furthermore, we demonstrate that the LLM-enabled domain information can be used to find good initializations in the search space for sampling-based SOTA methods to accelerate their convergence while preserving theoretical performance guarantees.
\end{abstract}

\section{Introduction}

The exponential increase in the volume and richness of available data has led to the widespread use of multi-way arrays, often represented as higher-order tensors. Tensor network decomposition methods aim to represent higher-order tensors in ``super-compressed'' formats through smaller-sized components, by effectively capturing cross-modal latent patterns and correlations. These methods have been applied across various fields, including machine learning \citep{td_1,td_3,td_2}, signal processing \citep{td_5}, computer vision \citep{td_6,td_4}, and quantum physics \citep{td_7,td_8}. The success of tensor network decomposition techniques is closely linked to their ability to mitigate the ``curse of dimensionality'', which is achieved by decomposing higher-order data into lower-order factors. 

However, tensor network (TN) practitioners face significant challenges related to choosing the most appropriate TN structure, the so called \textit{tensor network structure search (TN-SS)} problem, which has been proven to be NP-hard \citep{np_hard, tnale_cite}. TN-SS involves determining the optimal TN parameters, such as TN ranks, TN topology, and TN mode permutations \citep{tnls_cite}.

\begin{wrapfigure}{r}{0.45\columnwidth}
    \centering
    \includegraphics[width=1\linewidth]{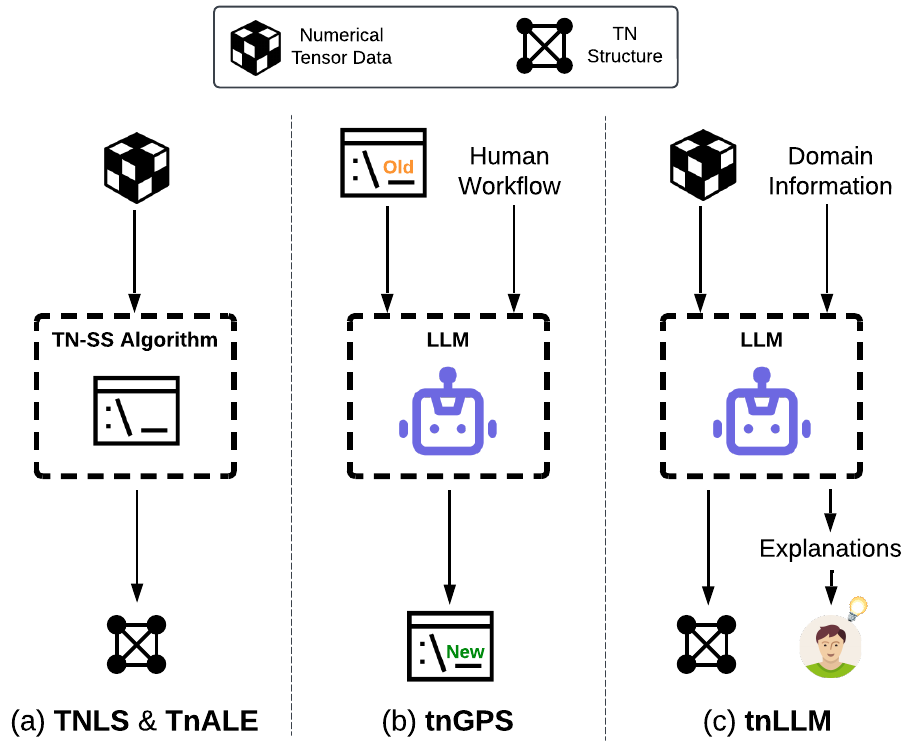}
    \caption{Comparison of SOTA tensor network structure search (TN-SS) methods. (a) TN-SS algorithms with theoretical guarantees: TNLS \citep{tnls_cite} \& TnALE \citep{tnale_cite}. (b) tnGPS \citep{zengtngps_cite}, a prompting pipeline which uses LLMs to generate new TN-SS algorithms. (c) tnLLM (ours), which uses domain information about the tensor data and LLM reasoning to solve the TN-SS problem and generate explanations for the identified TN structure.}
    \label{comparison_workflow}
\end{wrapfigure}

Existing TN-SS methods solve TN-SS as a purely numerical optimization problem and include approaches such as Bayesian inference \citep{bayes_1}, spectrum methods \citep{tn_order_2}, reinforcement learning \citep{rl_1}, discrete optimization \citep{tnls_cite,tnale_cite}, and continuous optimization \citep{svdins_tn}. To date, sampling-based methods \citep{tnga,tnls_cite,tnale_cite,zengtngps_cite}, whose workflows are illustrated in Figure \ref{comparison_workflow}a and \ref{comparison_workflow}b, have demonstrated the best performance in addressing the TN-SS problem. 

However, these methods require large number of evaluations (of the training and testing data), are prone to getting stuck in local minima, and lack transparency in their found structure-related parameters \citep{tnale_cite}. Critically, a large number of evaluations needed to optimize the objective function leads to a high computational cost. We hypothesize that these limitations arise from failing to exploit the rich domain information inherent in real-world tensors, such as the mode information.
To this end, we ask ourselves: \par 
\textit{How can we utilize the intrinsic domain information in tensor data to significantly reduce the number of evaluations required to solve the TN-SS problem, while providing domain-aware explanations for the identified TN solutions?} \par 

To address this question, we propose a domain-aware large language model (\underline{LLM})-guided \underline{TN}-SS framework, termed tnLLM. Within this framework, an LLM is utilized to initialize the TN structure based on domain information about the relationships between tensor modes. Then, the reasoning capabilities of the LLM are used to navigate the search space effectively, in order to achieve good optimization of the objective function with very few evaluations. 

Our proposed framework is found to achieve significant speed-ups over current state-of-the-art (SOTA) methods in terms of the number of evaluations, due to its ability to find good TN structure initializations through the use of domain information. Moreover, tnLLM generates practically meaningful explanations for the TN solutions, thus offering transparency in the identified structures. This is particularly beneficial for tensor practitioners who lack deep expertise in a specific data domain, as it enables them to both comprehend the interactions between tensor modes and trust the discovered TN structures. It also allows the identified structures to be verified by domain experts.\par 

To evaluate the effectiveness of tnLLM, we compare its performance against SOTA TN-SS algorithms on real-world tensor datasets of order-$3,4,\text{and }5 $. The experimental results demonstrate that tnLLM delivers performance comparable to current SOTA methods, while requiring significantly fewer function evaluations and providing domain-aware explanations for the identified TN structures. Moreover, we constructed a hybrid algorithm to combine the speed-up benefits of tnLLM with the theoretical guarantees of existing sampling-based approaches. The main contributions of this work are: 
\begin{itemize}
    \item We propose tnLLM, a novel domain-aware LLM-guided TN-SS framework, which achieves performance on par with SOTA methods while requiring much fewer evaluations; 
    \item To the best of our knowledge, our framework is the first to utilize domain information inherent in real-world tensor data to address the TN-SS problem. This enables the generation of domain-aware explanations that allows practitioners to verify the identified TN structures.
    
\end{itemize}
 
\begin{figure*}[!ht]
    \centering
    \includegraphics[width=0.95\linewidth]{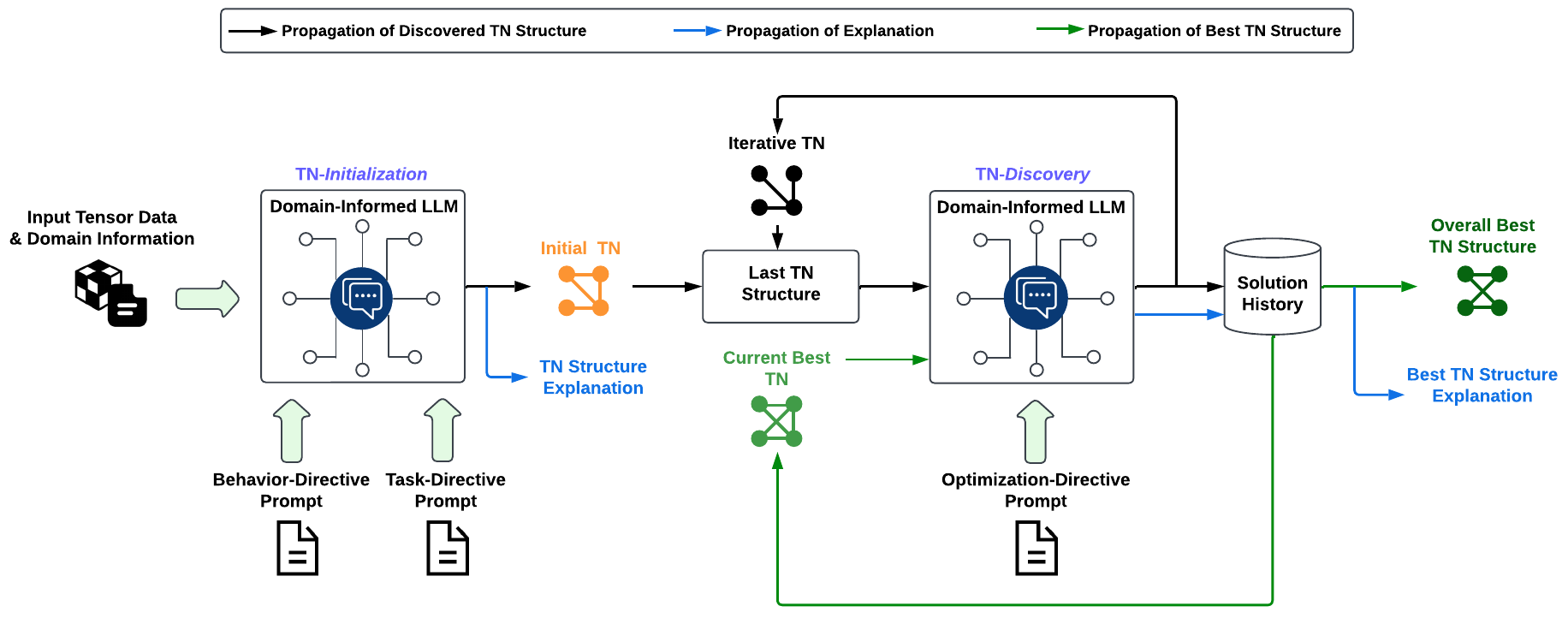}
    \caption{The workflow of the proposed tnLLM framework consists of two key stages commonly adopted by SOTA TN-SS methods \citep{tnls_cite, tnale_cite, zengtngps_cite}: \textit{TN-initialization} and \textit{\textit{TN-discovery}}. The proposed tnLLM efficiently utilises the domain information in tensor data and uses an LLM to guide these two stages.}
    \label{workflow}
\end{figure*}

\subsection{Related work} \label{related_work}
\paragraph{Tensor network structure search (TN-SS).} 
Compared to traditional tensor decompositions \citep{cpd, tkd, ttd, trd, twd}, which have pre-defined tensor network structures, the TN-SS problem focuses on finding custom tensor networks, which have been shown to achieve higher parameter efficiency and are an important paradigm of tensor decompositions \citep{tnls_cite,tnale_cite, tnga, zengtngps_cite}. Various approaches have been proposed to address the TN-SS problem, including Bayesian inference \citep{bayes_1}, spectrum methods \citep{tn_order_2}, reinforcement learning \citep{rl_1}, program synthesis \citep{tnss_program_synthesis}, and continuous optimization \citep{svdins_tn}. Sampling-based methods \citep{tnale_cite, tnls_cite, tnga, zengtngps_cite}, which fall under discrete optimization methods, have demonstrated superior performance compared to other approaches in addressing the TN-SS problem. Among these, TNLS \citep{tnls_cite} and TnALE \citep{tnale_cite} achieve state-of-the-art (SOTA) performance, with TnALE accelerating TNLS by slightly reducing the number of evaluations required to optimize the objective function. While both TNLS and TnALE adopt a “local-search” scheme, this acceleration is achieved by the alternating variable updates proposed in TnALE. However, these methods still face challenges, such as a large number of function evaluations required to converge, difficulty in balancing exploration and exploitation, and a lack of transparency in the found structures. \par 
More recently, tnGPS \citep{zengtngps_cite} has emerged as an approach that uses LLMs to generate sampling-based TN-SS algorithms, demonstrating performance comparable to SOTA methods. Despite being closest to our work, tnGPS focuses on generating TN-SS algorithms based on sampling-based heuristics and does not incorporate any domain information in the tensor data. As such, tnGPS suffers from the same limitations as existing sampling-based methods, including a lack of explainability in the identified structure-related parameters.\par 
To this end, our work proposes to harness the domain information in tensor data to solve the TN-SS problem. We achieve this by using the domain knowledge and the inherent reasoning capabilities of LLMs to \textit{directly} infer tensor network structures. Facilitated by our carefully designed prompting pipeline (see Section \ref{llm_info}), the proposed framework is shown not only to require significantly fewer evaluations, but also to produce domain-aware and verifiable solutions to the TN-SS problem.
\par
\paragraph{Reasoning with large language models.} The rise of transformer-based LLMs, pretrained on vast text corpora, has demonstrated a remarkable capacity for ``reasoning'' \citep{large_llm}. This reasoning ability is further enhanced when LLMs are guided by task-specific prompting strategies, such as chain-of thought \citep{reason_2,reason_4,reason_3}. 
These strategies enable LLMs to generate text effectively for tasks such as arithmetic reasoning and factual knowledge retrieval, leading to exceptional performance in complex question-answering and tasks requiring real-world knowledge \citep{qa_1,qa_2}. These findings suggest that, through pretraining on diverse textual data, LLMs encode rich knowledge about real-world relationships, which they can effectively leverage to perform various downstream tasks \citep{encoder_1}. In this work, we utilize the real-world domain knowledge and ``reasoning'' capabilities of LLMs to directly infer suitable TN structures.
\par

\section{Preliminaries}

\subsection{Tensor network structure search (TN-SS)}
We first provide the definition of TN-SS through its application in higher-order data decomposition. An optimal solution to the TN-SS problem aims to find the best trade-off between identifying the most compressed TN structure while preserving the expressivity of the TN \citep{li2021rank}. Let $\mathcal{X} \in \mathbb{R}^{I_1 \times I_2 \times \cdots \times I_N}$ be a non-zero order-$N$ tensor, with its Frobenius norm as $\|\mathcal{X}\|_F$.  Tensor networks can be represented using the graphical notation \citep{orus2014practical, ye2018tensor}, whereby a tensor network $G$ is represented using a set of $N$ vertices, $V$, and a set of edges, $E$, i.e., $G=(V,E)$. Each vertex represents a decomposed core tensor, and the closed edges between two core tensors are generalized higher-order matrix multiplications, termed tensor contractions \citep{td_3}. Closed edges have assigned \textit{TN-ranks}, $\mathbf{r} \in \mathbb{Z}_+^{E}$, which indicate the degree of connectivity between different pairs of connected vertices. Therefore, the properties of a TN structure can be fully expressed by the combination $(G,\mathbf{r})$.

Similar to SOTA TN-SS methods \citep{tnga, tnale_cite, tnls_cite, zengtngps_cite}, the discrete optimization problem of TN-SS is formalized as a minimization of the objective function, which is a linear sum of the complexity of the TN structure (e.g., compression rate) and the TN expressivity (e.g., approximation error), and is given by
\begin{equation} \label{loss}
    \min_{(G, \mathbf{r})} \ \ln \Bigg(\phi(G,\mathbf{r})  \\+ \frac{\lambda}{L} \min_{\{\mathcal{V}_{l,i}\}_{i=1}^N}   \sum_{l=1}^{L}     \frac{ \left\| \mathcal{X}_l - TNC\left(\{\mathcal{V}_{l,i}\}_{i=1}^N;(G,\mathbf{r})\right) \right\|_F}{\left\| \mathcal{X}_l \right\|_F} \Bigg)
\end{equation}
where $L$ represents the number of tensor samples in the dataset, and the first term $\phi(G,\mathbf{r})$ measures the TN structure complexity. The second term in Equation (\ref{loss}) measures the expressivity of the TN through the relative squared error (RSE) between the original tensors $\{\mathcal{X}_l\}_{l=1}^L$ and their TN approximations $\left\{TNC\left(\{\mathcal{V}_{l,i}\}_{i=1}^N;(G,\mathbf{r})\right)\right\}_{l=1}^L$, where $TNC(\cdot)$ stands for the tensor contraction operation of the entire tensor network. The pair $(G, \mathbf{r})$ characterizes how the vertices, $\{\mathcal{V}_{l,i}\}_{i=1}^N$, are contracted together to approximate the original tensor $\mathcal{X}_l$. The coefficient $\lambda$ is a positive non-zero scaling factor which balances the trade-off between model complexity and model expressivity. Note that \citep{tnga,ye2018tensor} pointed out that the TN-SS problem is conveniently equivalent to the TN rank search problem of a fully connected TN \citep{fctn}.

\begin{algorithm}[t]
\caption{\hspace{-1.75 mm} {\bf :} Sampling-based Algorithms for TN-SS \citep{tnls_cite, tnale_cite, zengtngps_cite}}
\small
\label{algo:SOTA_sampling}
\begin{algorithmic}[1]

\State \textbf{Initialize:}
\State $N_{Iter}$ \Comment{Maximum number of iterations}
\State $N_{Sample}$ \Comment{Number of samples per iteration}
\State $P \leftarrow [ ]$  \Comment{Historical TN structures}
\State $H \leftarrow [ ]$  \Comment{Discovered TN structures in each iter.}
\State $\mathcal{F}(\cdot) \leftarrow $ Equation (\ref{loss})  \Comment{Objective function}
\State $(G,\mathbf{r}) = \textit{Initial TN Structure}$ \Comment{TN-initialization}
\State
\State \textbf{Algorithm:}
\For{$n=1$ \textbf{to} $N_{Iter}$}
    \State $H \leftarrow $ $N_{Sample}$ TN structures sampled in the neighborhood of $(G,\mathbf{r})$. \Comment{TN-discovery} 
    \State $P \leftarrow P \cup H$ 
    \If{  $\exists (\hat G,\hat{\mathbf{r}}) \in P$ such that $\mathcal{F}(\hat G,\hat{\mathbf{r}})<\mathcal{F}( G,\mathbf{r})$}\
        \State $(G,\mathbf{r}) \leftarrow (\hat G,\hat{\mathbf{r}})$.
    \EndIf
    \If{Converged}
        \State \textbf{return} $(G,\mathbf{r})$
    \EndIf
\EndFor

\State \textbf{Output:} $(G,\mathbf{r})$

\end{algorithmic}
\end{algorithm}

\subsection{SOTA sampling-based algorithms for TN-SS}
Algorithm \ref{algo:SOTA_sampling} summarizes the current SOTA algorithms for TN-SS. They follow a three-step search process: \textit{\textit{TN-initialization}} $\rightarrow$ \textit{\textit{TN-discovery}} via sampling in the search neighborhood $\rightarrow$ Updating the center of the search neighborhood. Existing methods ignore the inherent domain information in real-world tensor data, which calls for the development of a framework that can effectively utilize domain knowledge in \textit{\textit{TN-initialization}} and \textit{\textit{TN-discovery}} to improve performance. \par 
To this end, our proposed framework introduces domain-aware LLM-guided \textit{\textit{TN-initialization}} and \textit{\textit{TN-discovery}}. By doing so, it addresses the limitations of existing methods, such as high computational costs caused by the large number of evaluations ($N_{Iter} \times N_{Sample}$) required and the tendency to get stuck in local minima due to difficulties in balancing the exploration-exploitation trade-off. Importantly, by incorporating domain information, our tnLLM framework provides practically meaningful explanations for the identified TN structures, a feature absent in current approaches.

\section{tnLLM: a domain-aware framework for solving TN-SS} \label{llm_info}
This section presents tnLLM, a domain-aware LLM-guided framework that efficiently solves the TN-SS problem with very few evaluations and verifiable solution explanations. We detail the role of each component in the prompting pipeline and the guidelines followed by each prompt. \par

\paragraph{Workflow of tnLLM.} Figure \ref{workflow} illustrates the workflow of the proposed tnLLM framework. The ``Behavior-directive'' prompt is used to supply the problem specification and guide the general-purpose LLM into a model tailored for solving the TN-SS problem. Inspired by current sampling-based SOTA algorithms, this domain-informed LLM is then employed for LLM-guided \textit{TN-initialization} and \textit{TN-exploration}. \par 
During the \textit{TN-initialization} stage, the ``Task-directive'' prompt is used to propose a strong initial TN structure by using the provided domain information.
In the \textit{TN-discovery} stage, the last and best-identified TN structures are used in conjunction with the ``Optimization-directive'' prompt to guide the LLM in navigating the search space effectively. The identified TN structures are fed into an objective function evaluation program to obtain the objective function values. This iterative process leads to the refinement of the TN structure and yields improved objective function values over successive evaluations.
\par

\paragraph{Prompting pipeline.} The interactions between the carefully designed user prompts and the LLM assistant follow an automated structured dialogue \citep{chat_1,chat_2}. 
This dialogue-based approach is intuitive for generating stepwise conversational reasoning and is particularly useful for iterative tasks such as the TN-SS problem. 
To further improve the ability of the LLM to perform complex reasoning, we employ the chain-of-thought prompting \citep{reason_2}. This prompting strategy breaks down complex tasks into a series of intermediate reasoning steps and thus enables the model to solve the problem by addressing smaller, simpler sub-tasks sequentially.  In the proposed tnLLM framework, three distinctly purposed prompts are employed, each serving a unique role. The full prompts are given in Appendix \ref{full_prompts}. \par
\begin{figure*}[!h]
    \centering
    \begin{subfigure}[t]{0.48\textwidth}
        \centering
        \includegraphics[width=\textwidth]{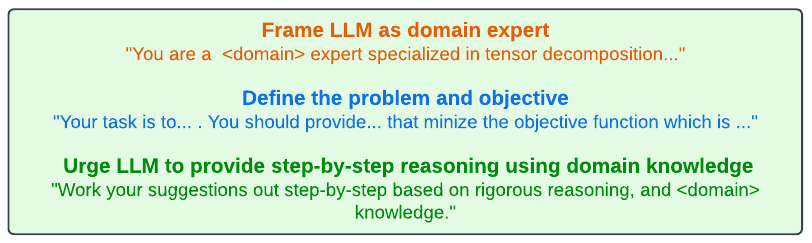}
        \par\vspace{0.5ex}
        \caption{“Behavior-directive” prompt}
        \label{system_prompt}
    \end{subfigure}%
    \hfill
    \begin{subfigure}[t]{0.48\textwidth}
        \centering
        \includegraphics[width=\textwidth]{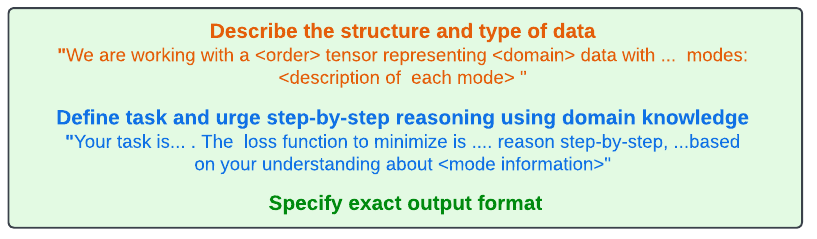}
        \par\vspace{0.5ex}
        \caption{“Task-directive” prompt}
        \label{initial_prompt}
    \end{subfigure}
    \caption{Structure and components of (a) the “Behavior-directive” prompt, which frames the LLM's role as a domain expert, and (b) the “Task-directive” prompt for LLM-guided \textit{TN-initialization.}}
    \label{behaviour_task_prompt}
\end{figure*}

 
\paragraph{\textbf{``Behavior-directive''} prompt.} This is the system message in the dialogue. Its primary objective is to establish the role of the LLM as an expert in the TN-SS problem (\textcolor{Orange}{orange} section of Figure \ref{system_prompt}) and guide its behavior throughout the dialogue (\textcolor{ForestGreen}{green} section of Figure \ref{system_prompt}). Additionally, it sets the context by outlining the TN-SS problem and the objective function (\textcolor{RoyalBlue}{blue} section of Figure \ref{system_prompt}). \par

\paragraph{\textbf{``Task-directive''} prompt.} This is the first non-system prompt supplied to the LLM and serves as the basis for LLM-guided \textit{TN-initialization}. It is designed to thoroughly explain the domain information about the tensor structure, including details such as the number of modes and the specific information about each mode (\textcolor{Orange}{orange} section of Figure \ref{initial_prompt}). Furthermore, it guides the model’s thought process and ensures that its responses align with the task requirements by instructing it to reason step-by-step and utilize domain knowledge (\textcolor{RoyalBlue}{blue} section of Figure \ref{initial_prompt}).

\par
\begin{wrapfigure}{r}{0.45\columnwidth}
    \centering
    \includegraphics[width=1\linewidth]{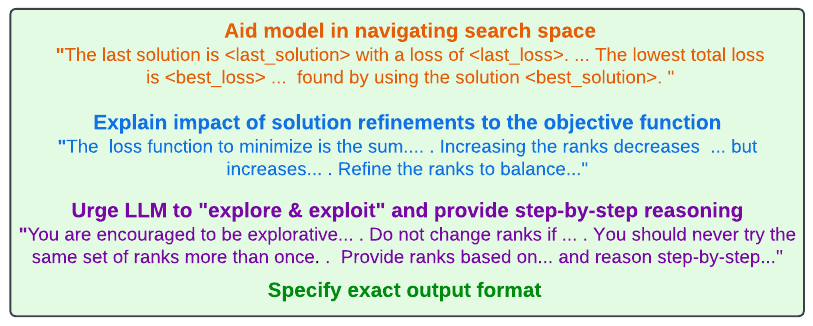}
    \caption{Structure and components of the ``Optimization-directive'' prompt used for LLM-guided \textit{TN-discovery}.}
    \label{iterative_prompt}
\end{wrapfigure}

 
\paragraph{\textbf{``Optimization-directive''} prompt.} This is the iterative prompt supplied to the LLM in all subsequent evaluations and serves as the basis for LLM-guided \textit{\textit{TN-discovery}}. Its purpose is to guide the LLM to efficiently navigate the search space by incorporating context from the best and previous evaluations (\textcolor{Orange}{orange} section of Figure \ref{iterative_prompt}). The prompt explains how refinements to the identified TN structure influence the objective function value and instructs the model to leverage step-by-step reasoning and domain knowledge to optimize the objective function (\textcolor{RoyalBlue}{blue} section of Figure  \ref{iterative_prompt}). At the same time, it encourages the LLM to both \textit{explore} and \textit{exploit} new solutions to reduce the likelihood of the method being stuck in local minima (\textcolor{violet}{purple} section of Figure  \ref{iterative_prompt}). \par 
\paragraph{Efficient and effective search space navigation.}  A critical factor in the performance of tnLLM, both in terms of evaluations and objective function value, is the incorporation of memory (\textcolor{Orange}{orange}  section of Figure \ref{iterative_prompt}) and the encouragement of search space "exploration and exploitation" (\textcolor{violet}{purple} section of Figure \ref{iterative_prompt}) in the ``Optimization-directive'' prompt. By including the best and most recent TN solutions, along with their respective objective function values and the domain information about mode interactions, the LLM gains a better understanding of the search space and can make more informed decisions. Additionally, by encouraging the model to explore new solutions when necessary and exploit good solutions already found assists the optimization process, by reducing the possibility of the model getting stuck in local minima, a common caveat of sampling-based methods. \par

\begin{wrapfigure}{r}{0.45\columnwidth}
    \centering
    \includegraphics[width=1\linewidth]{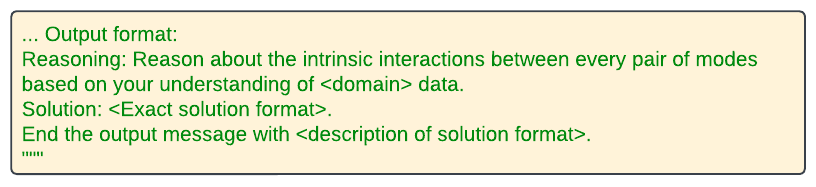}
     \caption{The prompt used to specify the LLM output format.}
     \label{fig:exact_output_prompt}
\end{wrapfigure}

\paragraph{Output format specification.} As observed in Figures \ref{initial_prompt} and \ref{iterative_prompt}, in addition to the goal-oriented components of each prompt, an essential feature (highlighted in \textcolor{ForestGreen}{green}) of both the ``Task-directive'' and ``Optimization-directive'' prompts is the specification of the exact output format of the TN-SS solution. This includes defining both the sequence in which the LLM should present its response and the precise format of the TN-SS solution. To meet these requirements, the prompt shown in Figure \ref{fig:exact_output_prompt} was developed through experimentation and a trial-and-error process with the LLM. This step serves to eliminate ``hallucinations'' in the LLM output and enables a \textit{fully automated} prompting pipeline without the need for any human intervention during the entire iterative optimization process.
\begin{remark}
    Both tnGPS and tnLLM use LLMs to solve the TN-SS problem, however, the two approaches are fundamentally different. tnGPS uses the LLM to generate new \textit{sampling-based TN-SS algorithms} by reviewing existing algorithms, causing the discovered methods to inherit similar limitations. In contrast, tnLLM uses the LLM to \textit{directly infer TN structures} to solve the TN-SS problem by incorporating the rich domain information in real-world tensor data. In turn, this allows tnLLM to provide explanations for the TN solutions, thereby adding transparency to the identified TN structures.
\end{remark}
\section{Experimental results} \label{results}
\label{sec:exp}
In this section, the performance of tnLLM was evaluated against the SOTA sampling-based methods in tensor decomposition tasks in terms of the number of evaluations and objective function value. Our results demonstrate that tnLLM achieves comparable objective function values while delivering significant speed-ups. Moreover, we demonstrate its ability to generate domain-consistent explanations for the identified TN structures. Finally, an ablation study was conducted to evaluate the effectiveness of domain information and assess the framework's robustness across different LLMs.  \par 

\paragraph{Data preparation.} Given the flexibility of our method to handle tensor data of any order, we evaluated its performance on three types of tensor data with varying sizes and dimensionalities across different domains. In particular, we tested on datasets of order-$3$ RGB images 
and order-$4$ RGB videos.
We also curated a completely new order-$5$ financial time-series dataset of $142$ tensors to ensure that it was not included in the training data of LLMs.
This time-series tensor dataset, to the best of our knowledge, is the largest dataset in terms of number of samples ever considered in the TN-SS problem. All entries were standardized to values in $[0,1]$, with $80\%$ of each dataset used for training, and the remaining $20\%$ for testing. More details about the data can be found in Appendix \ref{sec:data}. \par 

\paragraph{Settings of tnLLM.} In Equation (\ref{loss}), we set $\lambda=10$ and used the same compression ratio function, $\phi$, as in previous sampling-based methods \citep{tnga}, defined as the ratio between the number of parameters in the compressed TN format and the original tensor. The maximum number of evaluations was set to $500$ for the images and videos datasets, and $250$ for the financial time-series dataset. An early stopping criterion with a patience of 5 was applied. For all experiments, the LLM model GPT-4o (\texttt{gpt-4o-2024-08-06}) \citep{HelloGPT4o} was used, with temperature set at 0.2. \par 

\paragraph{Implementation details.} In all experiments, we also implemented the three SOTA sampling-based TN-SS algorithms, namely TNLS \citep{tnls_cite}, TnALE \citep{tnale_cite} and tnGPS \citep{zengtngps_cite}, and accelerated them with GPUs. Since the vanilla TNLS is designed to search only for the permutation of a TN, we extended it to fit the settings of TN-SS. For fair comparisons, all baseline methods were evaluated using the same objective function and maximum number of evaluations, with one evaluation defined as a single pass through the entire training and testing dataset. The full implementation details are provided in Appendix \ref{imp_details}.\par

\paragraph{Numerical results.} We ran tnLLM five times and examined the mean and standard deviation for both the objective function values and the number of evaluations to demonstrate its \textit{robustness}. Observe from Table \ref{tab:comparison_sec_1} that tnLLM achieved performance on par with SOTA sampling-based algorithms in both training and test objectives across all data types, as measured by the objective function. Importantly, tnLLM minimizes the objective function with significantly fewer evaluations, requiring up to $78 \times$ fewer evaluations than TNLS, $41 \times$ fewer than TnALE, and $110\times$ fewer than tnGPS. Consequently, tnLLM achieves runtime reductions of up to 98.3\% compared to TNLS, 97.7\% compared to TnALE, and 98.1\% compared to tnGPS, even after accounting for LLM inference.
The full runtime comparisons are provided in Appendix \ref{runtime}. \par

\begin{table*}[t]
\centering
\scriptsize
\caption{Performance comparison across different datasets. The values on the left give the lowest training and corresponding testing objective function values. The values in [square brackets] give the number of evaluations required to first achieve the best training objective function value. For robustness assessment of tnLLM, we report the average and standard deviation of both the objective function value and the number of evaluations across 5 independent runs. For both metrics, a lower value is better. The best values are denoted in bold. The second best values are underlined.}
\label{tab:comparison_sec_1}
\begin{tabular}{ l l  cc  cc  cc  cc }
\toprule
\textbf{Data Type} & 
  & \multicolumn{2}{c}{\textbf{TNLS}} 
  & \multicolumn{2}{c}{\textbf{TnALE}} 
  & \multicolumn{2}{c}{\textbf{tnGPS}} 
  & \multicolumn{2}{c}{\textbf{tnLLM (Ours)}} \\

\midrule
\multirow{2}{*}{\textbf{Images}} 
  & \textbf{Train} 
      & -0.66  &  \multirow{2}{*}{[114]}
      & -0.65  &  \multirow{2}{*}{[\underline{81}]}
      & -0.66  &  \multirow{2}{*}{[438]}  
      & -0.63$\pm0.01$  &  \multirow{2}{*}{[\textbf{4.0}$\pm$1.9]} \\ 
  & \textbf{Test}  
      & \underline{-0.47}  
      &        & -0.46  
      &        & -0.44  
      &        & \textbf{-0.48}$\pm$0.05
      &        \\
\midrule

\multirow{2}{*}{\textbf{Videos}} 
  & \textbf{Train} 
      & -1.64  &  \multirow{2}{*}{[484]}  
      & -1.66  &  \multirow{2}{*}{[254] } 
      & -1.65  &  \multirow{2}{*}{[\underline{175}]}
      & -1.63$\pm0.01$  &  \multirow{2}{*}{[\textbf{6.2}$\pm$3.5]} \\ 
  & \textbf{Test}  
      & \textbf{-1.72}  
      &        & \textbf{-1.72}  
      &        & \underline{-1.71}  
      &        & -1.70$\pm$0.02
      &        \\
\midrule

\multirow{2}{*}{\textbf{Time-series}} 
  & \textbf{Train} 
      & -0.45  &  \multirow{2}{*}{[218]}  
      & -0.47  &  \multirow{2}{*}{[177]}  
      & -0.39 &  \multirow{2}{*}{[\underline{38}] }  
      & -0.42$\pm0.02$  &  \multirow{2}{*}{[\textbf{5.6}$\pm$4.4]}\\ 
  & \textbf{Test}  
      & \underline{-0.43}  
      &        & \textbf{-0.47}  
      &        &    -0.40    &       
      & -0.41$\pm$0.02
      &        \\
\bottomrule
\end{tabular}
\end{table*}

\begin{figure}[!h]
    \centering
    \includegraphics[width=0.93\linewidth]{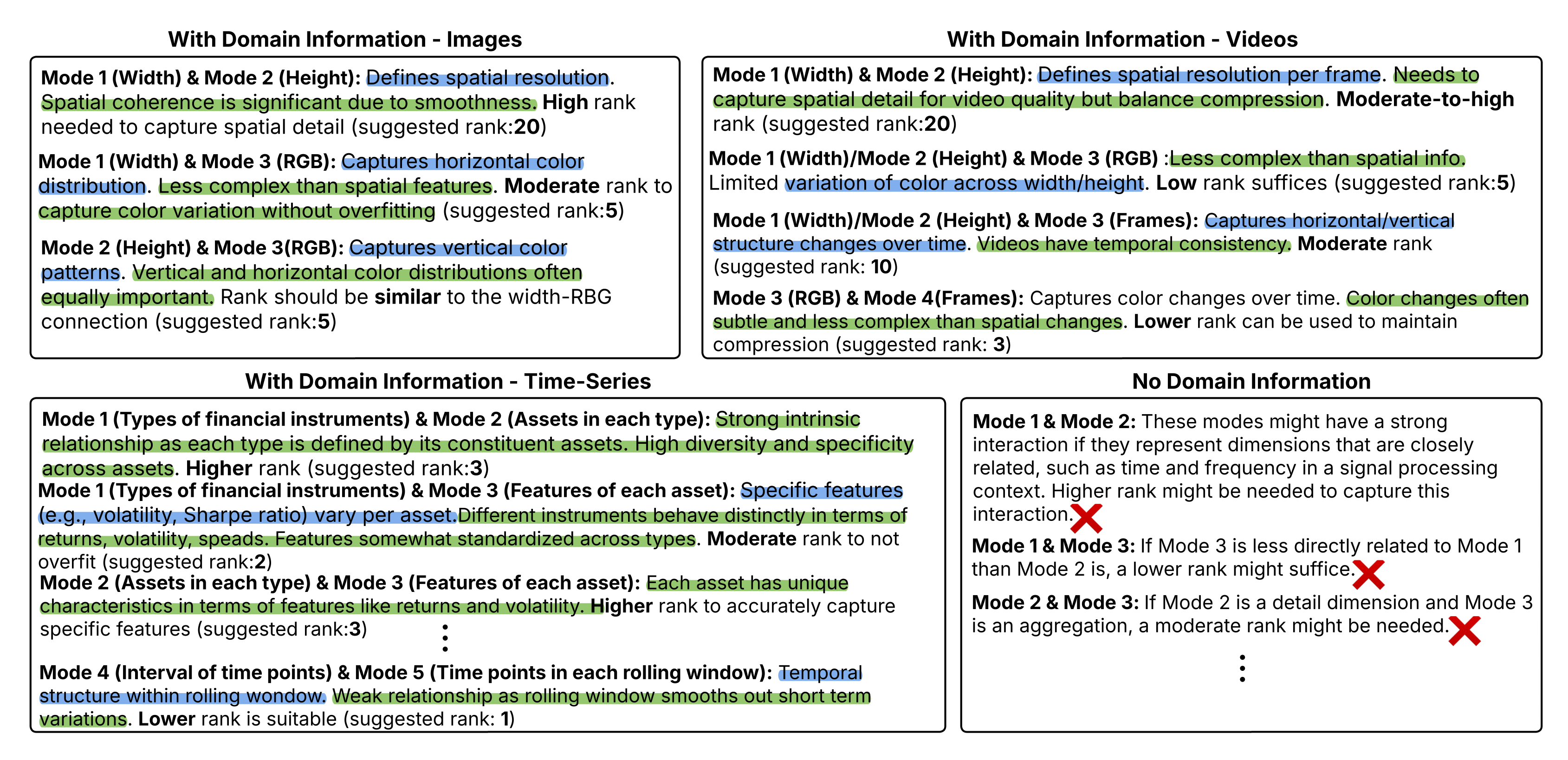}
    \caption{Summarized generated explanations of tnLLM for the initialized TN structure, with and without the incorporation of domain information in the prompts.}
    \label{interp_result}
\end{figure}

\paragraph{Domain-aware explanations.} To assess the domain-relevance of the identified TNs, we manually verified the explanations provided for the identified TN structures in all three datasets tested. The explanations of the tnLLM model for the initialized TN structure, with and without domain knowledge, are provided in Figure \ref{interp_result}. Observe that without domain information, although the model is inherently capable of reasoning, its lack of domain information about the modes in the real-world tensors causes it to default to random assumptions, likely influenced by its pre-training data, such as assuming a mode as resembling time or frequency. As a result, the explanations are entirely incorrect from a practical perspective and are therefore ineffective in solving TN-SS. \par 

In contrast, incorporating domain information enables the model to identify relationships between modes (highlighted in \textcolor{RoyalBlue}{blue}) and, more importantly, relate them to real-world principles in vision and finance  (highlighted in \textcolor{ForestGreen}{green}). The model also shows consistent reasoning in its rank suggestions: when describing a rank as low, medium, or high, it quantitatively selects a value that is coherent relative to other ranks within the same tensor. Moreover, it adjusts these values based on comparisons across different mode pairs, for example by assigning similar ranks to the Height–RGB and Width–RGB modes in the images dataset, and a lower rank to the RGB–Frames mode in the videos dataset. This demonstrates the model's understanding of tensor mode relationships based on domain information and supports the validity of the identified TN structure. Consequently, tnLLM offers explanations that are practically useful in helping domain experts to verify and trust the discovered structures. Summarized explanations across three runs for all datasets are provided in Appendix \ref{full_interp_results}. \par

\subsection{Accelerating SOTA methods with tnLLM} \label{accelerating_sota}
In the first part of Section \ref{results}, we have demonstrated that tnLLM achieves comparable objective function values compared to SOTA sampling-based algorithms,  while generating domain-aware explanations for the identified TN structures and significantly reducing the required number of function evaluations. However, due to the black-box nature of LLMs, no theoretical analysis can be provided for the evaluation efficiency of tnLLM, in contrast to TNLS and TnALE. \par 
At the same time, while sampling-based methods follow a ``local-search'' scheme within a neighborhood, there is no guarantee on the minimum number of evaluations required to find a ``good'' neighborhood. Moreover, poor initialization significantly increases the number of evaluations needed. To address this, we construct a hybrid algorithm that first runs tnLLM for 10 evaluations to leverage its domain 
knowledge and reasoning capabilities in order to identify a strong initialization point in the ``global-search'' stage. Sampling-based methods then perform ``local-search'' in the identified neighborhood, combining the speed-up benefits of tnLLM with their theoretical guarantees.

\begin{figure}[!h]
    \centering
    \includegraphics[width=0.95\linewidth]{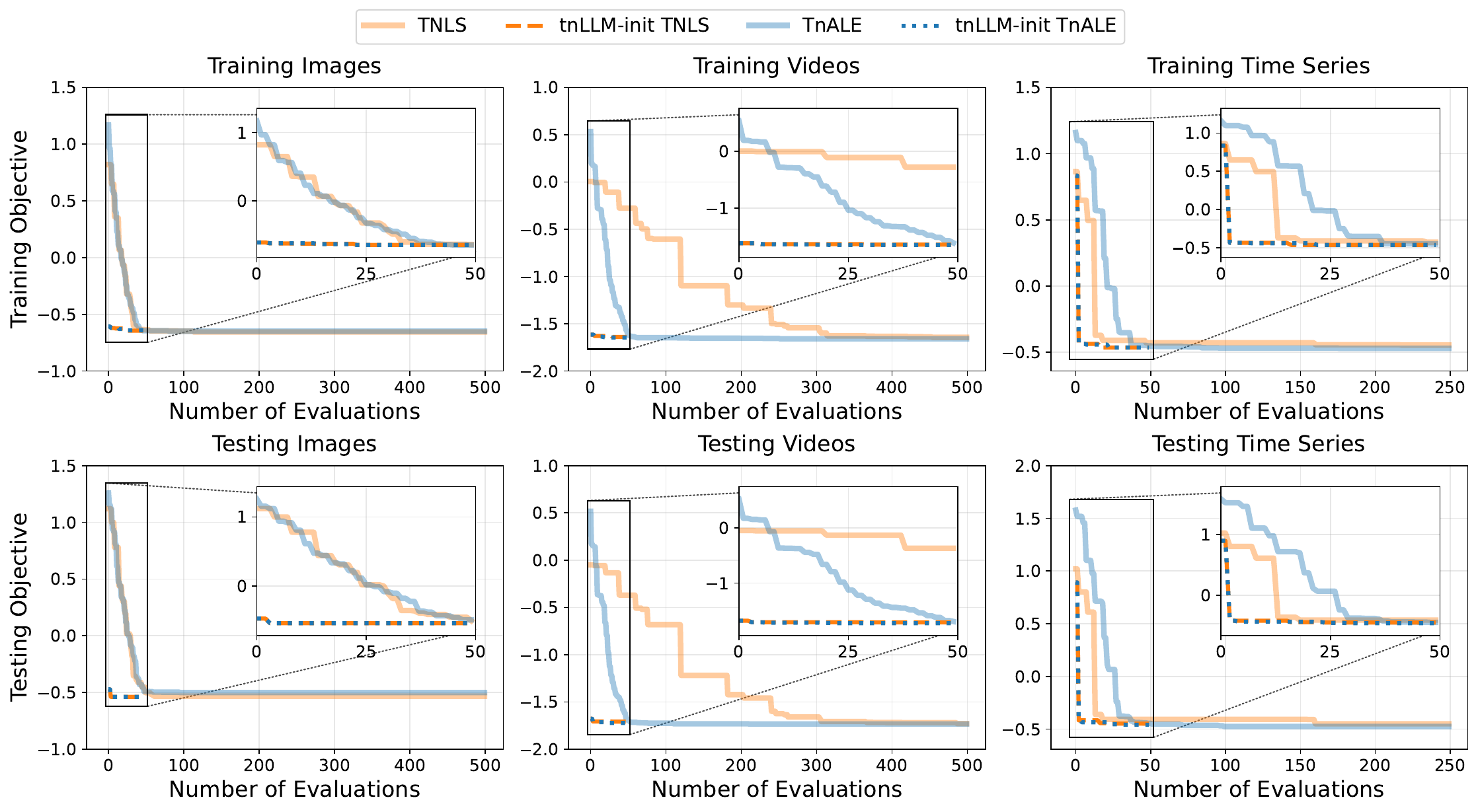}
    \caption{Experimental results across all 3 datasets show that both TNLS and TnALE are significantly accelerated with an initial center of search neighborhood found by tnLLM. We ran TNLS and TnALE with tnLLM initialization in all 3 datasets over 50 evaluations, which \textit{include} the plotted 10 evaluations performed by tnLLM. Vanilla TNLS and TnALE performed 500 evaluations for both the images and videos datasets and 250 evaluations for the financial time-series dataset if not converged.}
    \label{init_comparison_plot}
    
\end{figure}
\par 
\paragraph{Numerical results.} Figure \ref{init_comparison_plot} demonstrates the performance difference of TNLS and TnALE with and without a tnLLM-initialized structure. The ``global-search'' capability of tnLLM significantly sped up the iterative minimization process. During ``local-search'', both TNLS and TnALE further improved upon the structures found in ``global-search'' by tnLLM. Overall, the tnLLM-initialized algorithms achieved nearly identical objective function values with up to $23\times$ fewer evaluations compared to vanilla TNLS, and up to $13\times$ fewer evaluations compared to vanilla TnALE. The full numerical performance comparison is provided in Appendix \ref{sec:perf_tnllm_init}.  

\subsection{Ablation study} \label{ablation_study}

\paragraph{Removal of domain information.} To assess the effectiveness of injecting domain information into the TN-SS problem, we removed the carefully designed structured prompts of the tnLLM framework, equivalent to using the LLM to solve the TN-SS problem without any priors on the domain information. Without domain information, the LLM generates a poorly initialized TN structure that is 80.3\% worse in terms of objective function value, due to its lack of information about the different modes, and requires $10.4 \times$ more evaluations to converge. Moreover, as the model defaults to random assumptions about the modes, as illustrated in Figure \ref{interp_result}, it was found to produce uniformly connected TN structures throughout the minimization process. This restricts the search space to a much smaller set of solutions that are not domain-meaningful from a practical perspective.

\paragraph{Selection of LLM models.} To examine how the choice of LLM affects the performance of tnLLM, we compared our baseline model, GPT-4o (\texttt{gpt-4o-2024-08-06}), against GPT-4.5 (\texttt{gpt-4.5-preview-2025-02-27}), GPT4o-mini (\texttt{gpt-4o-mini-2024-07-18}), GPT-3.5 (\texttt{gpt-3.5-turbo-1106}) and the open-source DeepSeek V3 (\texttt{DeepSeek-V3-0324}) model. Observe from Table \ref{abl_llm} that due to the carefully designed prompts and the structure of the overall framework, tnLLM is robust to the choice of the LLM used. While there are small variations in the achieved objective function values, the overall performance remained consistent across all three datasets in both training and testing sets. Furthermore, the number of evaluations required by all LLMs falls within the mean ± standard deviation range reported in Table~\ref{tab:comparison_sec_1}. It is important to note that, despite the consistent performance, weaker models and in particular GPT-3.5 occasionally misinterpreted the optimization objective, mistakenly assuming that higher values were preferable. 

\begin{table}[h]
\scriptsize
\caption{Ablation study of using different LLMs within the proposed framework. For different LLMs, we report the best training objective values achieved and their corresponding testing objective values.}
\centering
\begin{tabular}{lcccccccccc}
\toprule
\textbf{Dataset} & \multicolumn{2}{c}{\textbf{GPT-4o (Baseline)}} & \multicolumn{2}{c}{\textbf{GPT-4.5}} & \multicolumn{2}{c}{\textbf{GPT-4o-mini}} & \multicolumn{2}{c}{\textbf{GPT-3.5}}
& \multicolumn{2}{c}{\textbf{DeepSeek-V3}} \\
 & Train & Test & Train & Test & Train & Test & Train & Test & Train & Test \\
\midrule
Images      & -0.63 & -0.48 & -0.62 & -0.54 & -0.61 & -0.48 & -0.64 & -0.52 & -0.62 & -0.47 \\
Videos      & -1.63 & -1.70 & -1.62 & -1.65 & -1.62 & -1.69 & -1.63 & -1.70 & -1.63 & -1.70 \\
Time-series & -0.42 & -0.41 & -0.40 & -0.41 & -0.41 & -0.40 & -0.42 & -0.40 & -0.40 & -0.39\\
\bottomrule
\end{tabular}

\label{abl_llm}
\end{table}

\section{Conclusion}
We have introduced tnLLM, a domain-aware LLM-guided TN-SS framework for directly infering TN structures. This has been achieved by utilizing the rich domain information in real-world tensor data and the inherent reasoning capabilities of LLMs. Experimental results have demonstrated that tnLLM achieves performance comparable to current SOTA algorithms, while requiring significantly fewer function evaluations. Notably, by incorporating domain information, our framework is the first to mitigate the black-box nature of the identified TN structures in TN-SS through generating domain-relevant solution explanations. Furthermore, we have shown that tnLLM can be used to accelerate SOTA sampling-based algorithms while preserving their theoretical guarantees. 

\bibliography{references.bib}

\begin{thebibliography}{38}
\providecommand{\natexlab}[1]{#1}
\providecommand{\url}[1]{\texttt{#1}}
\expandafter\ifx\csname urlstyle\endcsname\relax
  \providecommand{\doi}[1]{doi: #1}\else
  \providecommand{\doi}{doi: \begingroup \urlstyle{rm}\Url}\fi

\bibitem[Chen et~al.(2024)Chen, Lu, and Zhang]{tn_order_2}
Ziang Chen, Jianfeng Lu, and Anru~R. Zhang.
\newblock One-dimensional tensor network recovery, 2024.

\bibitem[Choi et~al.(2022)Choi, Cundy, Srivastava, and Ermon]{encoder_1}
Kristy Choi, Chris Cundy, Sanjari Srivastava, and Stefano Ermon.
\newblock {LMP}riors: Pre-trained language models as task-specific priors.
\newblock In \emph{NeurIPS 2022 Foundation Models for Decision Making Workshop}, 2022.

\bibitem[Cichocki et~al.(2016)Cichocki, Lee, Oseledets, Phan, Zhao, and Mandic]{td_3}
Andrzej Cichocki, Namgil Lee, Ivan Oseledets, Anh~Huy Phan, Qibin Zhao, and Danilo~P Mandic.
\newblock Tensor networks for dimensionality reduction and large-scale optimization: Part 1 low-rank tensor decompositions.
\newblock \emph{Foundations and Trends® in Machine Learning}, 9\penalty0 (4-5):\penalty0 249--429, 2016.

\bibitem[Felser et~al.(2021)Felser, Trenti, Sestini, Gianelle, Zuliani, Lucchesi, and Montangero]{td_8}
Tobias Felser, Michele Trenti, Lorenzo Sestini, Alessandro Gianelle, Davide Zuliani, Donatella Lucchesi, and Simone Montangero.
\newblock Quantum-inspired machine learning on high-energy physics data.
\newblock \emph{npj Quantum Information}, 7\penalty0 (1):\penalty0 111, 2021.

\bibitem[Guo et~al.(2025)Guo, Deshpande, Kiedrowski, Wang, and Gorodetsky]{tnss_program_synthesis}
Zheng Guo, Aditya Deshpande, Brian Kiedrowski, Xinyu Wang, and Alex Gorodetsky.
\newblock Tensor network structure search using program synthesis.
\newblock \emph{arXiv preprint arXiv:2502.02711}, 2025.

\bibitem[Hillar \& Lim(2013)Hillar and Lim]{np_hard}
Christopher~J. Hillar and Lek-Heng Lim.
\newblock {Most tensor problems are NP-hard}.
\newblock \emph{Journal of the ACM}, 60\penalty0 (6):\penalty0 1--39, 2013.

\bibitem[Hitchcock(1927)]{cpd}
Frank~L Hitchcock.
\newblock The expression of a tensor or a polyadic as a sum of products.
\newblock \emph{Journal of Mathematics and Physics}, 6\penalty0 (1-4):\penalty0 164--189, 1927.

\bibitem[Kojima et~al.(2024)Kojima, Gu, Reid, Matsuo, and Iwasawa]{reason_3}
Takeshi Kojima, Shixiang~Shane Gu, Machel Reid, Yutaka Matsuo, and Yusuke Iwasawa.
\newblock Large language models are zero-shot reasoners.
\newblock In \emph{Proceedings of the 36th International Conference on Neural Information Processing Systems}, 2024.

\bibitem[Li \& Sun(2020)Li and Sun]{tnga}
Chao Li and Zhun Sun.
\newblock Evolutionary topology search for tensor network decomposition.
\newblock In \emph{Proceedings of the 37th International Conference on Machine Learning}, volume 119 of \emph{Proceedings of Machine Learning Research}, pp.\  5947--5957. PMLR, 13--18 Jul 2020.

\bibitem[Li \& Zhao(2021)Li and Zhao]{li2021rank}
Chao Li and Qibin Zhao.
\newblock Is rank minimization of the essence to learn tensor network structure.
\newblock In \emph{Second Workshop on Quantum Tensor Networks in Machine Learning (QTNML), Neurips}, volume~3, 2021.

\bibitem[Li et~al.(2022)Li, Zeng, Tao, and Zhao]{tnls_cite}
Chao Li, Junhua Zeng, Zerui Tao, and Qibin Zhao.
\newblock Permutation search of tensor network structures via local sampling.
\newblock In \emph{Proceedings of the 39th International Conference on Machine Learning}, pp.\  13106--13124, Jul 2022.

\bibitem[Li et~al.(2023)Li, Zeng, Li, Caiafa, and Zhao]{tnale_cite}
Chao Li, Junhua Zeng, Chunmei Li, Cesar Caiafa, and Qibin Zhao.
\newblock {Alternating local enumeration (TnALE): solving tensor network structure search with fewer evaluations}.
\newblock In \emph{Proceedings of the 40th International Conference on Machine Learning}, ICML'23. JMLR.org, 2023.

\bibitem[Liévin et~al.(2023)Liévin, Hother, Motzfeldt, and Winther]{qa_1}
Valentin Liévin, Christoffer~Egeberg Hother, Andreas~Geert Motzfeldt, and Ole Winther.
\newblock Can large language models reason about medical questions?
\newblock \emph{arXiv}, 2023.

\bibitem[Malik(2021)]{td_2}
Osman~Asif Malik.
\newblock More efficient sampling for tensor decomposition.
\newblock \emph{ArXiv}, abs/2110.07631, 2021.

\bibitem[Meirom et~al.(2022)Meirom, Maron, Mannor, and Chechik]{rl_1}
Eli Meirom, Haggai Maron, Shie Mannor, and Gal Chechik.
\newblock Optimizing tensor network contraction using reinforcement learning.
\newblock In Kamalika Chaudhuri, Stefanie Jegelka, Le~Song, Csaba Szepesvari, Gang Niu, and Sivan Sabato (eds.), \emph{Proceedings of the 39th International Conference on Machine Learning}, pp.\  15278--15292. PMLR, 17--23 Jul 2022.

\bibitem[OpenAI(2023)]{chat_1}
OpenAI.
\newblock {GPT-4 Technical Report}.
\newblock Technical report, OpenAI, 2023.

\bibitem[OpenAI(2024)]{HelloGPT4o}
OpenAI.
\newblock {Hello GPT-4o}.
\newblock \url{https://openai.com/index/hello-gpt-4o/}, 2024.

\bibitem[Or{\'u}s(2014)]{orus2014practical}
Rom{\'a}n Or{\'u}s.
\newblock A practical introduction to tensor networks: Matrix product states and projected entangled pair states.
\newblock \emph{Annals of Physics}, 349:\penalty0 117--158, 2014.

\bibitem[Orús(2019)]{td_7}
Rom{\'a}n Orús.
\newblock Tensor networks for complex quantum systems.
\newblock \emph{Nature Reviews Physics}, 1\penalty0 (9):\penalty0 538--550, 2019.

\bibitem[Oseledets(2011)]{ttd}
Ivan~V Oseledets.
\newblock Tensor-train decomposition.
\newblock \emph{SIAM Journal on Scientific Computing}, 33\penalty0 (5):\penalty0 2295--2317, 2011.

\bibitem[Shakeri \& Zhang(2019)Shakeri and Zhang]{td_6}
Moein Shakeri and Hong Zhang.
\newblock Moving object detection under discontinuous change in illumination using tensor low-rank and invariant sparse decomposition.
\newblock In \emph{Proceedings of the IEEE/CVF Conference on Computer Vision and Pattern Recognition (CVPR)}, pp.\  7221--7230. IEEE, 2019.

\bibitem[Singhal et~al.(2023)Singhal, Azizi, Tu, et~al.]{qa_2}
Karan Singhal, Shekoofeh Azizi, Tinsu Tu, et~al.
\newblock Large language models encode clinical knowledge.
\newblock \emph{Nature}, 620:\penalty0 172--180, 2023.
\newblock \doi{10.1038/s41586-023-06291-2}.

\bibitem[Suzgun et~al.(2023)Suzgun, Scales, Sch{\"a}rli, Gehrmann, Tay, Chung, Chowdhery, Le, Chi, Zhou, and Wei]{reason_4}
Mirac Suzgun, Nathan Scales, Nathanael Sch{\"a}rli, Sebastian Gehrmann, Yi~Tay, Hyung~Won Chung, Aakanksha Chowdhery, Quoc Le, Ed~Chi, Denny Zhou, and Jason Wei.
\newblock Challenging {BIG}-bench tasks and whether chain-of-thought can solve them.
\newblock In \emph{Findings of the Association for Computational Linguistics: ACL 2023}. Association for Computational Linguistics, July 2023.

\bibitem[Touvron et~al.(2023)Touvron, Lavril, Izacard, Martinet, Lachaux, Lacroix, Rozi{\`e}re, Goyal, Hambro, Azhar, et~al.]{chat_2}
Hugo Touvron, Thibaut Lavril, Gautier Izacard, Xavier Martinet, Marie-Anne Lachaux, Timoth{\'e}e Lacroix, Baptiste Rozi{\`e}re, Naman Goyal, Eric Hambro, Faisal Azhar, et~al.
\newblock Llama: Open and efficient foundation language models.
\newblock \emph{arXiv preprint arXiv:2302.13971}, 2023.

\bibitem[Tucker(1966)]{tkd}
Ledyard~R Tucker.
\newblock Some mathematical notes on three-mode factor analysis.
\newblock \emph{Psychometrika}, 31\penalty0 (3):\penalty0 279--311, 1966.

\bibitem[Wei et~al.(2022{\natexlab{a}})Wei, Tay, Bommasani, Raffel, Zoph, Borgeaud, Yogatama, Bosma, Zhou, Metzler, Chi, Hashimoto, Vinyals, Liang, Dean, and Fedus]{large_llm}
Jason Wei, Yi~Tay, Rishi Bommasani, Colin Raffel, Barret Zoph, Sebastian Borgeaud, Dani Yogatama, Maarten Bosma, Denny Zhou, Donald Metzler, Ed~H. Chi, Tatsunori Hashimoto, Oriol Vinyals, Percy Liang, Jeff Dean, and William Fedus.
\newblock Emergent abilities of large language models.
\newblock \emph{Transactions on Machine Learning Research}, 2022{\natexlab{a}}.
\newblock ISSN 2835-8856.

\bibitem[Wei et~al.(2022{\natexlab{b}})Wei, Wang, Schuurmans, Bosma, brian ichter, Xia, Chi, Le, and Zhou]{reason_2}
Jason Wei, Xuezhi Wang, Dale Schuurmans, Maarten Bosma, brian ichter, Fei Xia, Ed~H. Chi, Quoc~V Le, and Denny Zhou.
\newblock {Chain of thought prompting elicits reasoning in large language models}.
\newblock In \emph{Advances in Neural Information Processing Systems}, 2022{\natexlab{b}}.

\bibitem[Wu et~al.(2022)Wu, Huang, Deng, Dou, and Meng]{twd}
Zhong-Cheng Wu, Ting-Zhu Huang, Liang-Jian Deng, Hong-Xia Dou, and Deyu Meng.
\newblock Tensor wheel decomposition and its tensor completion application.
\newblock \emph{Advances in Neural Information Processing Systems}, 35:\penalty0 27008--27020, 2022.

\bibitem[Yamamoto et~al.(2022)Yamamoto, Hontani, Imakura, and Yokota]{td_4}
Ryuki Yamamoto, Hidekata Hontani, Akira Imakura, and Tatsuya Yokota.
\newblock Fast algorithm for low-rank tensor completion in delay-embedded space.
\newblock In \emph{Proceedings of the IEEE/CVF Conference on Computer Vision and Pattern Recognition (CVPR)}, pp.\  2048--2056. IEEE, 2022.

\bibitem[Ye \& Lim(2018)Ye and Lim]{ye2018tensor}
Ke~Ye and Lek-Heng Lim.
\newblock Tensor network ranks.
\newblock \emph{arXiv preprint arXiv:1801.02662}, 2018.

\bibitem[Yokota et~al.(2018)Yokota, Erem, Guler, Warfield, and Hontani]{yokota2018missing}
Tatsuya Yokota, Burak Erem, Seyhmus Guler, Simon~K Warfield, and Hidekata Hontani.
\newblock Missing slice recovery for tensors using a low-rank model in embedded space.
\newblock In \emph{Proceedings of the IEEE Conference on Computer Vision and Pattern Recognition}, pp.\  8251--8259, 2018.

\bibitem[Zeng et~al.(2024{\natexlab{a}})Zeng, Li, Sun, Zhao, and Zhou]{zengtngps_cite}
Junhua Zeng, Chao Li, Zhun Sun, Qibin Zhao, and Guoxu Zhou.
\newblock {tnGPS: Discovering unknown tensor network structure search algorithms via large language models (LLMs)}.
\newblock In \emph{Proceedings of the Forty-first International Conference on Machine Learning}, 2024{\natexlab{a}}.

\bibitem[Zeng et~al.(2024{\natexlab{b}})Zeng, Zhou, Qiu, Li, and Zhao]{bayes_1}
Junhua Zeng, Guoxu Zhou, Yuning Qiu, Chao Li, and Qibin Zhao.
\newblock Bayesian tensor network structure search and its application to tensor completion.
\newblock \emph{Neural Networks}, 175\penalty0 (C), July 2024{\natexlab{b}}.

\bibitem[Zhao et~al.(2016)Zhao, Zhou, Xie, Zhang, and Cichocki]{trd}
Qibin Zhao, Guoxu Zhou, Shengli Xie, Liqing Zhang, and Andrzej Cichocki.
\newblock Tensor ring decomposition.
\newblock \emph{arXiv preprint arXiv:1606.05535}, 2016.

\bibitem[Zhe et~al.(2015)Zhe, Xu, Chu, Qi, and Park]{td_1}
Shandian Zhe, Zenglin Xu, Xinqi Chu, Yuan Qi, and Youngja Park.
\newblock {Scalable nonparametric multiway data analysis}.
\newblock In \emph{Proceedings of the Eighteenth International Conference on Artificial Intelligence and Statistics}, pp.\  1125--1134. PMLR, 2015.

\bibitem[Zheng et~al.(2023)Zheng, Zhao, Zheng, Lin, Zhuang, and Huang]{td_5}
Wen-Jie Zheng, Xi-Le Zhao, Yu-Bang Zheng, Jie Lin, Lina Zhuang, and Ting-Zhu Huang.
\newblock Spatial-spectral-temporal connective tensor network decomposition for thick cloud removal.
\newblock \emph{ISPRS Journal of Photogrammetry and Remote Sensing}, 199:\penalty0 182--194, 2023.

\bibitem[Zheng et~al.(2021)Zheng, Huang, Zhao, Zhao, and Jiang]{fctn}
Yu-Bang Zheng, Ting-Zhu Huang, Xi-Le Zhao, Qibin Zhao, and Tai-Xiang Jiang.
\newblock Fully-connected tensor network decomposition and its application to higher-order tensor completion.
\newblock \emph{Proceedings of the AAAI Conference on Artificial Intelligence}, 35\penalty0 (12):\penalty0 11071--11078, May 2021.

\bibitem[Zheng et~al.(2024)Zheng, Zhao, Zeng, Li, Zhao, Li, and Huang]{svdins_tn}
Yu-Bang Zheng, Xi-Le Zhao, Junhua Zeng, Chao Li, Qibin Zhao, Heng-Chao Li, and Ting-Zhu Huang.
\newblock {SVDinsTN: A tensor network paradigm for efficient structure search from regularized modeling perspective}.
\newblock In \emph{Proceedings of the IEEE/CVF Conference on Computer Vision and Pattern Recognition}, 2024.

\end{thebibliography}
\bibliographystyle{iclr2026_conference}

\newpage
\appendix
\section{Limitations and future work} \label{apendix_limitations}
Despite the good performance, the proposed domain-aware TN-SS framework currently lacks theoretical guarantees for its evaluation efficiency due to the black-box nature of LLMs. This could be mitigated by developing domain-aware TN-SS algorithms based on continuous optimization. Also, as mentioned in the ablation study, weaker LLM models, such as GPT-3.5, were found to be more prone to misinterpret the optimization objectives. Therefore, improving the reasoning consistency of smaller LLM models promises to further improve the efficiency of the framework. Furthermore, prompt learning is an active area of research in LLMs, which can be used to potentially enhance the tnLLM framework in future works. 

\section{Full prompts of tnLLM framework} \label{full_prompts}
\begin{figure}[!h]
    \centering
    \includegraphics[width=1\textwidth]{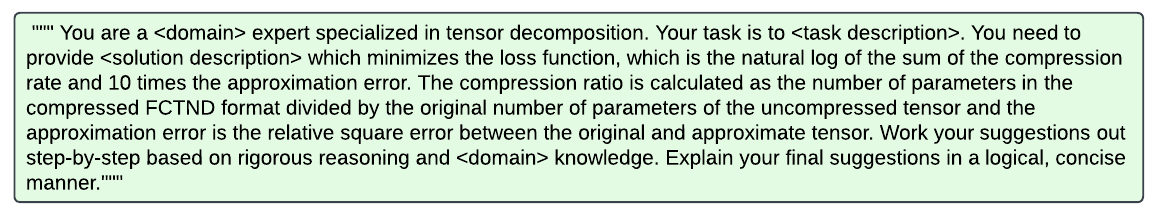}
    \caption{The full ``Behavior-directive'' prompt.}
    \label{full_syst_msg}
\end{figure}

\begin{figure}[!h]
    \centering
    \includegraphics[width=1\textwidth]{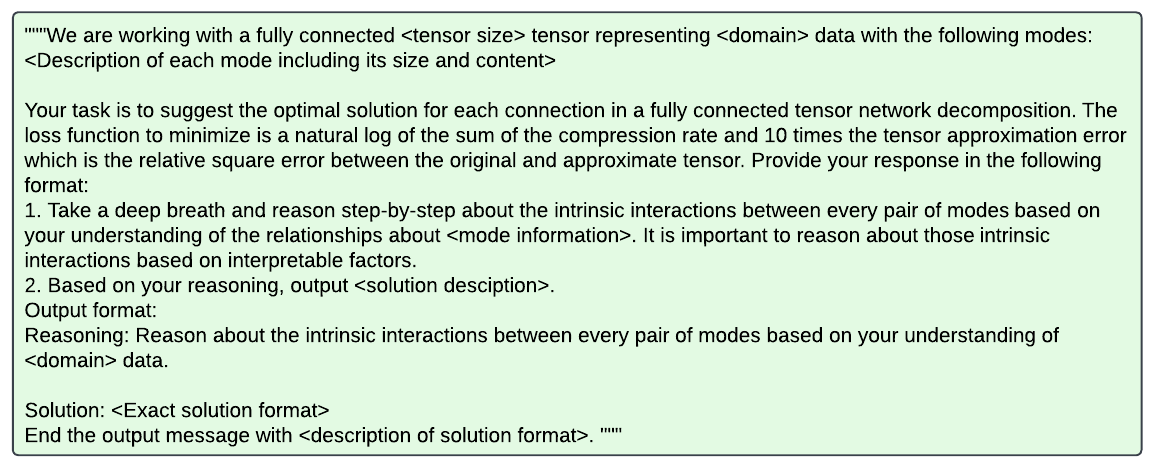}
    \caption{The full ``Task-directive'' prompt.}
    \label{full_init_prompt}
\end{figure}

\begin{figure}[!h]
    \centering
    \includegraphics[width=1\textwidth]{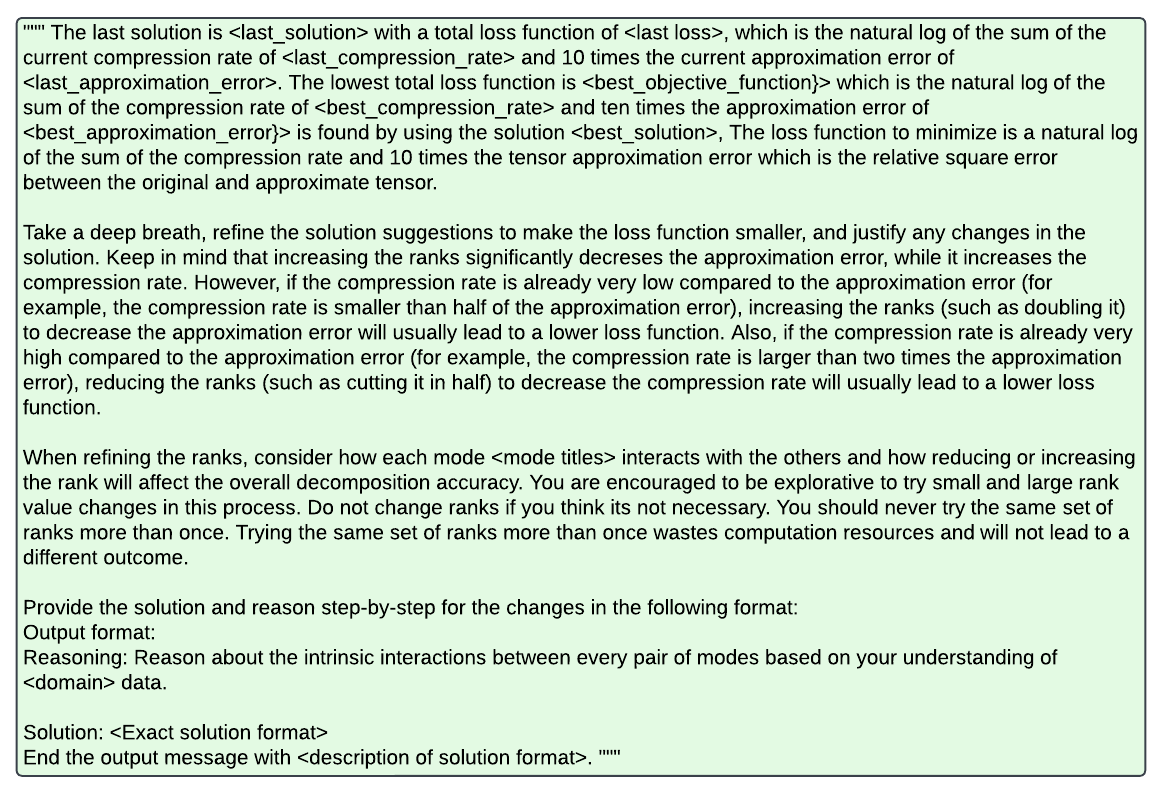}
    \caption{The full ``Optimization-directive'' prompt.}
    \label{full_iter_prompt}
\end{figure}

The full ``Behavior-directive'', ``Task-directive'' and ``Optimization-directive'' prompts used in the tnLLM framework are illustrated in Figures \ref{full_syst_msg}, \ref{full_init_prompt}, \ref{full_iter_prompt}, respectively.
\section{Data} \label{sec:data}

We constructed a custom financial time series dataset with 142 temporally ordered fifth-order tensors, denoted as $\{\mathcal{X}_n\}_{n=1}^{142} \in \mathbb{R}^{3 \times 6 \times 3 \times 4 \times 5}$, each representing a rolling window produced via multi-way delay embedding through the temporal direction \citep{yokota2018missing}. This leads to the value selection process of $10$ ranks. To the best of our knowledge, this is the largest tensor dataset in terms of number of samples ever considered in the TN-SS problem. The first $80\%$ of these tensors were used as the training data, while the remaining $20\%$ with non-overlapping entries with the training data were used for testing. The modes of each time series tensor correspond to:
\begin{itemize} \item \textbf{Mode 1}: Types of financial instruments. They are equity indices, commodities, and currency swaps. \item \textbf{Mode 2}: Assets within each type of financial instrument. For equity indices, these are Hang Seng, Nikkei 225, S\&P 500, EURO STOXX 50, FTSE 100, and Shanghai Composite Index. For commodities, these are Brent Crude, Copper, Natural gas, Comex gold, Soybeans, and Wheat. For currency swaps, these are HKD/USD, JPY/USD, CHF/USD, EUR/USD, CNY/USD, and GBP/USD. \item \textbf{Mode 3}: Features of each asset. These are average adjusted closing price log return, average relative price min-max, and average high-low spread. \item \textbf{Mode 4}: Interval of time points on which we calculate the average features. There are 4 intervals -- 1 day, 5 days, 10 days, and 15 days. \item \textbf{Mode 5}: Time points within each rolling window of length 5. \end{itemize}

\begin{figure}[!h]
    \centering
    \includegraphics[width=1\textwidth]{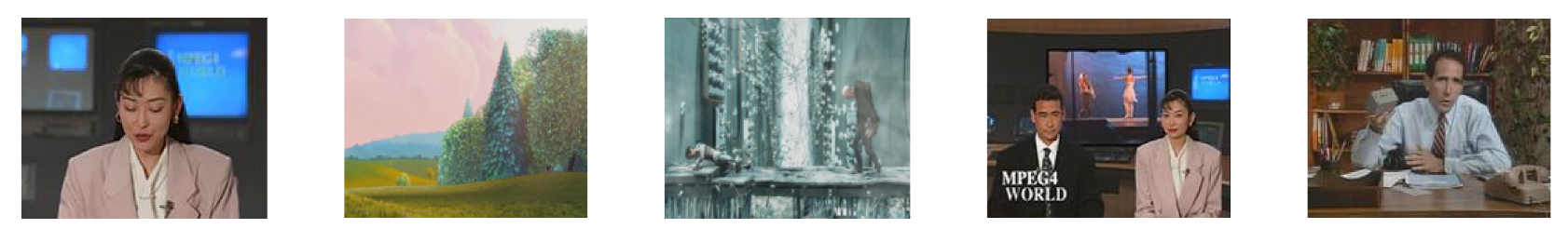}
    \caption{Image samples used in the experiments.}
    \label{fig:image_data}
\end{figure}

The images and videos datasets are fetched from \textit{http://trace.eas.asu.edu/yuv/}. Figure \ref{fig:image_data} illustrates 5 example image samples, $\{\mathcal{X}_n\}_{n=1}^{5} \in \mathbb{R}^{144 \times 176 \times 3}$, used in the experiments. The first $80\%$ of these tensors were used as the training data, while the remaining $20\%$ for testing. The modes of each image tensor correspond to:
\begin{itemize} \item \textbf{Mode 1}: Height of the image. These are the vertical pixel indexes. \item \textbf{Mode 2}: Width of the image. These are the horizontal pixel indexes. \item \textbf{Mode 3}: RGB channels. These are the red, green, and blue channels. \end{itemize}

\begin{figure}[!h]
    \centering
    \includegraphics[width=1\textwidth]{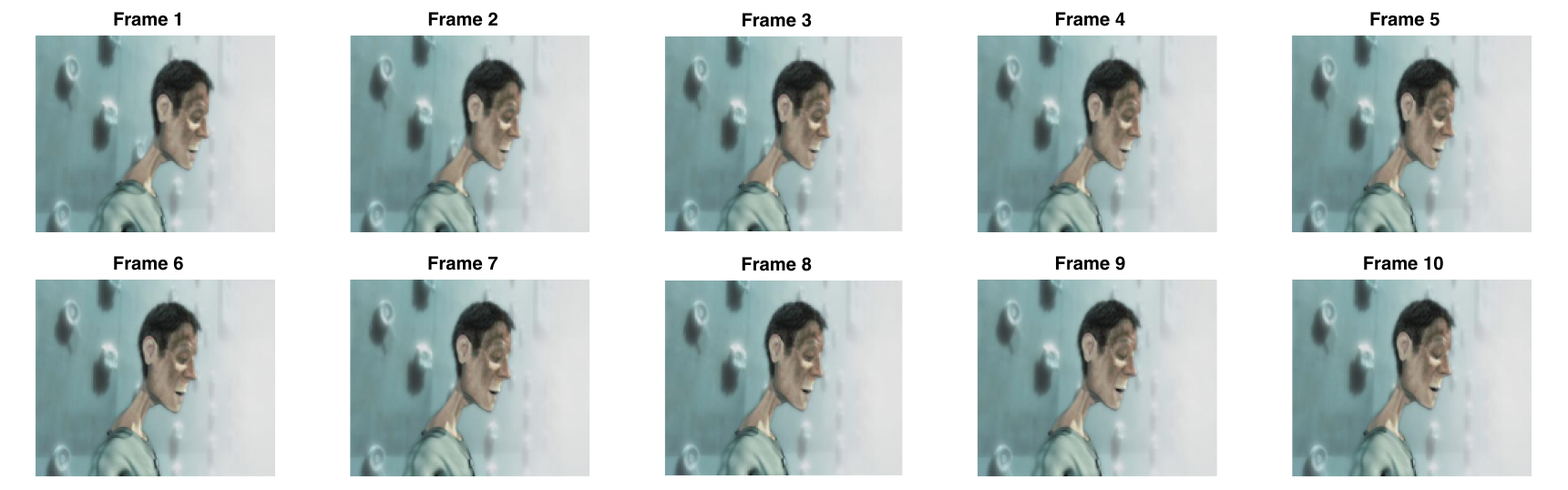}
    \caption{One example sample of the Video data used in the experiments.}
    \label{fig:video_data}
\end{figure}

Video dataset is produced from performing the multi-way delay embedding \citep{td_4} through the temporal mode in a video to produce 5 samples, $\{\mathcal{X}_n\}_{n=1}^{5} \in \mathbb{R}^{144 \times 176 \times 3 \times 10}$. Figure \ref{fig:video_data} shows an example sample of the videos dataset at different frames. The modes of each video tensor correspond to:
\begin{itemize} \item \textbf{Mode 1}: Height of the image. These are the vertical pixel indexes. \item \textbf{Mode 2}: Width of the image. These are the horizontal pixel indexes. \item \textbf{Mode 3}: RGB channels. These are the red, green, and blue channels. \item \textbf{Mode 4}: Frames in a video. These are the index of the frames in a video. \end{itemize}

\section{Implementation details} \label{imp_details} All baseline models are initialized as a ``fully-disconnected'' graph. The TN search template is set to a complete graph. For TNLS, we set the number of samples (evaluations), $\#Sample$, in each sampling step to $4$ in the images dataset, $60$ in the videos dataset, and $10$ in the time series dataset. For TnALE, we set $D=1$, $r=1$. For tnGPS, we selected the best-performing algorithm discovered (``Ho-2'') and adopted the following hyperparameters as suggested in their paper: code upper bound of 10, mutation rate of 0.1, crossover rate of 0.6, selection pressure of 1.5, elitism enabled, diversity factor of 0.05, variance decay of 0.98, minimum variance of 0.1, tournament size factor of 0.2, elite diversity boost of 2.0, random individual chance of 0.05, and a maximum mutation of 3. All models were run for $500$ evaluations in both the images and videos dataset, and $250$ evaluations in the time series dataset if not converged. An internal server with NVIDIA RTX A6000 GPU, an AMD Ryzen Threadripper PRO 5955WX with 16 cores, and 256GB of RAM was used.

\section{Runtime Comparisons} \label{runtime}
Table \ref{tab:runtime} reports the total average runtime in seconds to first achieve the best training objective function value for all models across the three datasets of Table \ref{tab:comparison_sec_1}. For tnLLM, this also includes the LLM inference time. The proposed tnLLM achieves comparable performance to SOTA methods while reducing runtime by up to 98.3\% compared to TNLS, 97.7\% compared to TnALE, and 98.1\% compared to tnGPS.

\begin{table}[h!]
\centering
\caption{Runtime comparisons across three datasets. The values are the total average runtime in seconds to first achieve the best training objective function value. A lower value is better. The best values are denoted in bold. The second best values are underlined.\\}
\begin{tabular}{lcccc}
\toprule
\textbf{Data Type} & \textbf{TNLS} & \textbf{TnALE} & \textbf{tnGPS} & \textbf{tnLLM (Ours)} \\
\midrule
Images & 2,122s & \underline{1,659s} & 8,025s & \textbf{151s} \\
Videos & 17,690s & \underline{15,298s} & 20,461s & \textbf{1,984s} \\
Time-series & 110,945s & 80,700s & \underline{17,287s} & \textbf{1,866s} \\
\bottomrule
\end{tabular}
\label{tab:runtime}
\end{table}

\section{Summarized generated explanations of tnLLM for the initialized TN structure} \label{full_interp_results}
The full set of summarized LLM generated explanations over three independent runs for the images, videos, and time-series datasets are provided in Tables~\ref{llm_interp_image}, \ref{llm_interp_video}, and \ref{llm_interp_time}, respectively. The suggested ranks are verified to be consistent with the domain-informed reasoning provided by tnLLM. Also, the domain-relevant explanations are robust across multiple runs.

\begin{table*}[!h]
\centering
\scriptsize
\caption{Summarized generated explanations of tnLLM for the initialized TN structure in the Images dataset over 3 independent runs.\\}
\label{llm_interp_image}
\renewcommand{\arraystretch}{1.3}
\begin{tabularx}{\textwidth}{>{\raggedright\arraybackslash}p{4cm} 
                                >{\raggedright\arraybackslash}X 
                                >{\raggedright\arraybackslash}X 
                                >{\raggedright\arraybackslash}X}
\toprule
\textbf{Mode Pair} & \textbf{Run 1} & \textbf{Run 2} & \textbf{Run 3} \\
\midrule
\textbf{Mode 1 \& Mode 2} \newline (Width $\leftrightarrow$ Height)
& \begin{itemize}[left=0pt]
    \item Defines spatial resolution
    \item Spatial coherence is significant due to smoothness
    \item \textbf{High rank} needed to capture spatial detail (suggested rank: 20)
\end{itemize}
& \begin{itemize}[left=0pt]
    \item Spatial dimensions are highly correlated
    \item Captures patterns and structures
    \item \textbf{High rank} to capture spatial complexity (suggested rank: 20)
\end{itemize}
& \begin{itemize}[left=0pt]
    \item Spatial resolution key for details and patterns
    \item \textbf{High rank} crucial to capture more spatial features to maintain image quality
    \item \textbf{High rank} (suggested rank: 20)
\end{itemize}
\\
\midrule
\textbf{Mode 1 \& Mode 3} \newline (Width $\leftrightarrow$ RGB)
& \begin{itemize}[left=0pt]
    \item Captures horizontal color distribution
    \item Less complex than spatial features
    \item \textbf{Moderate rank} to capture color variation without overfitting (suggested rank: 5)
\end{itemize}
& \begin{itemize}[left=0pt]
    \item Horizontal color changes are smooth
    \item Color variations less complex as object maintains consistent color across width
    \item \textbf{Moderate rank} (suggested rank: 5)
\end{itemize}
& \begin{itemize}[left=0pt]
    \item Related through color distribution along width
    \item Color variations less complex than spatial variation
    \item \textbf{Moderate rank} (suggested rank: 5)
\end{itemize}
\\
\midrule
\textbf{Mode 2 \& Mode 3} \newline (Height $\leftrightarrow$ RGB)
& \begin{itemize}[left=0pt]
    \item Captures vertical color patterns
    \item Vertical and horizontal color distributions often equally important
    \item Rank should be \textbf{similar} to the width–RGB connection (suggested rank: 5)
\end{itemize}
& \begin{itemize}[left=0pt]
    \item Captures vertical color variation
    \item Vertical structures (e.g., stripes) may add complexity
    \item Rank might be \textbf{slightly higher} than width–channel rank (suggested rank: 8)
\end{itemize}
& \begin{itemize}[left=0pt]
    \item Complexity of color variation along height and width typically similar
    \item \textbf{Moderate rank} is also appropriate (suggested rank: 5)
\end{itemize}
\\
\bottomrule
\end{tabularx}
\end{table*}

\begin{table*}[!h]
\centering
\scriptsize
\caption{Summarized generated explanations of tnLLM for the initialized TN structure in the Videos dataset over 3 independent runs.\\}
\label{llm_interp_video}
\renewcommand{\arraystretch}{1.3}
\begin{tabularx}{\textwidth}{>{\raggedright\arraybackslash}p{4.3cm} 
                                >{\raggedright\arraybackslash}X 
                                >{\raggedright\arraybackslash}X 
                                >{\raggedright\arraybackslash}X}
\toprule

\textbf{Mode Pair} & \textbf{Run 1} & \textbf{Run 2} & \textbf{Run 3} \\
\midrule
\textbf{Mode 1 \& Mode 2} \newline (Width $\leftrightarrow$ Height)
& \begin{itemize}[left=0pt]
    \item Defines spatial resolution per frame
    \item Needs to capture spatial detail for video quality but balance with compression
    \item \textbf{Moderate-to-high rank} (suggested rank: 20)
\end{itemize}
& \begin{itemize}[left=0pt]
    \item Define spatial resolution per frame
    \item Higher rank captures more complex spatial patterns and correlations
    \item \textbf{Moderate-to-high rank} to balance detail capture and compression (suggested rank: 20)
\end{itemize}
& \begin{itemize}[left=0pt]
    \item Define spatial resolution per frame
    \item Spatial resolution is a significant aspect of video quality
    \item \textbf{Moderate-to-high rank} to capture more spatial details across frames (suggested rank: 20)
\end{itemize}
\\
\midrule
\textbf{Mode 1 \& Mode 3, Mode 2 \& Mode 3} \newline (Width/Height $\leftrightarrow$ RGB)
& \begin{itemize}[left=0pt]
    \item Less complex than spatial info
    \item Limited variation of color across width/height
    \item \textbf{Low rank} suffices (suggested rank: 5)
\end{itemize}
& \begin{itemize}[left=0pt]
    \item Less directly correlated than spatial dimensions
    \item Certain horizontal/vertical patterns prominent in specific color channels
    \item \textbf{Low rank} suffices (suggested rank: 5)
\end{itemize}
& \begin{itemize}[left=0pt]
    \item Captures color variations horizontally/vertically
    \item Color variations can be significant in videos with rich color content
    \item \textbf{Moderate-to-low rank} (suggested rank: 5)
\end{itemize}
\\
\midrule
\textbf{Mode 1 \& Mode 4, Mode 2 \& Mode 4} \newline (Width/Height $\leftrightarrow$ Frames)
& \begin{itemize}[left=0pt]
    \item Captures horizontal/vertical structure changes over time
    \item Videos have temporal consistency
    \item \textbf{Moderate rank} (suggested rank: 10)
\end{itemize}
& \begin{itemize}[left=0pt]
    \item Related through motion and changes across frames
    \item \textbf{Moderate rank} to capture temporal changes across width/height of the video (suggested rank: 10)
\end{itemize}
& \begin{itemize}[left=0pt]
    \item Captures temporal variations across the horizontal/vertical dimension
    \item Temporal changes crucial for motion representation
    \item \textbf{Moderate rank} (suggested rank: 10)
\end{itemize}
\\
\midrule
\textbf{Mode 3 \& Mode 4} \newline (RGB $\leftrightarrow$ Frames)
& \begin{itemize}[left=0pt]
    \item Captures color changes over time
    \item Color changes often subtle and less complex than spatial changes
    \item \textbf{Lower rank} can be used to maintain compression (suggested rank: 3)
\end{itemize}
& \begin{itemize}[left=0pt]
    \item Captures color changes over time
    \item Crucial for capturing dynamic color variations and transitions
    \item \textbf{Low rank} to model temporal color variation (suggested rank: 5)
\end{itemize}
& \begin{itemize}[left=0pt]
    \item Captures temporal color changes
    \item Essential for representing dynamic scenes
    \item \textbf{Moderate-to-low rank} necessary to capture these changes (suggested rank: 5)
\end{itemize}
\\
\bottomrule
\end{tabularx}
\end{table*}

\begin{table*}[!h]
\centering
\scriptsize 
\caption{Summarized generated explanations of tnLLM for the initialized TN structure in the Time-series dataset over 3 independent runs.\\}

\label{llm_interp_time}
\renewcommand{\arraystretch}{1.3}
\begin{tabularx}{\textwidth}{>{\raggedright\arraybackslash}p{4.5cm} 
                                >{\raggedright\arraybackslash}X 
                                >{\raggedright\arraybackslash}X 
                                >{\raggedright\arraybackslash}X}
\toprule
\textbf{Mode Pair} & \textbf{Run 1} & \textbf{Run 2} & \textbf{Run 3} \\
\midrule
\textbf{Mode 1 \& Mode 2} \newline (Types of financial instruments $\leftrightarrow$ Assets within each type)
& \begin{itemize}[left=0pt]
    \item Strong intrinsic relationship
    \item Different assets within a type exhibit correlated behaviors due to shared market factors
    \item \textbf{Higher rank} (suggested rank: 3)
\end{itemize}
& \begin{itemize}[left=0pt]
    \item Strong relationship as each type is defined by its constituent assets
    \item High diversity and specificity across assets
    \item \textbf{Higher rank} (suggested rank: 3)
\end{itemize}
& \begin{itemize}[left=0pt]
    \item Strong connection as each instrument has specific assets
    \item Rank should reflect the diversity and specificity of assets within each type
    \item \textbf{Higher rank} (suggested rank: 3)
\end{itemize}
\\
\midrule
\textbf{Mode 1 \& Mode 3} \newline (Types of financial instruments $\leftrightarrow$ Features of each asset)
& \begin{itemize}[left=0pt]
    \item Features influenced by instrument type
    \item For example, volatility of commodities might differ from that of equity indexes
    \item \textbf{Moderate rank} (suggested rank: 2)
\end{itemize}
& \begin{itemize}[left=0pt]
    \item Different instruments behave distinctly (returns, volatility, spreads)
    \item Features somewhat standardized across types
    \item \textbf{Moderate rank} to not overfit (suggested rank: 2)
\end{itemize}
& \begin{itemize}[left=0pt]
    \item Instruments behave differently in terms of features
    \item However, features themselves are common across all types
    \item \textbf{Moderate rank} to capture variability (suggested rank: 2)
\end{itemize}
\\
\midrule
\textbf{Mode 2 \& Mode 3} \newline (Assets within each type $\leftrightarrow$ Features of each asset)
& \begin{itemize}[left=0pt]
    \item Features directly derived from assets indicating strong relationship
    \item Different assets have distinct feature profiles
    \item \textbf{Higher rank} (suggested rank: 3)
\end{itemize}
& \begin{itemize}[left=0pt]
    \item Each asset has unique characteristics in terms of returns, volatility, and spreads
    \item \textbf{Higher rank} to accurately capture specific features (suggested rank: 3)
\end{itemize}
& \begin{itemize}[left=0pt]
    \item Each asset has unique characteristics in terms of features like log returns and spreads
    \item \textbf{Higher rank} needed for this detailed interaction (suggested rank: 3)
\end{itemize}
\\
\midrule
\textbf{Mode 4 \& Mode 5} \newline (Interval of time points $\leftrightarrow$ Time points in each rolling window)
& \begin{itemize}[left=0pt]
    \item Inherently linked as rolling window defines the context for the time points
    \item \textbf{Moderate rank} (suggested rank: 2)
\end{itemize}
& \begin{itemize}[left=0pt]
    \item Weak relationship as rolling window smooths out short-term variations
    \item \textbf{Lower rank} is suitable (suggested rank: 1)
\end{itemize}
& \begin{itemize}[left=0pt]
    \item Intrinsic connection as time points are nested within the intervals
    \item \textbf{Moderate rank} to capture this structure (suggested rank: 2)
\end{itemize}
\\
\bottomrule
\end{tabularx}
\end{table*}

\section{Performance comparison for vanilla and tnLLM-initialized TNLS and TnALE}
\label{sec:perf_tnllm_init}
Table \ref{tab:comparison_sec_2} illustrates the detailed performance of vanilla and tnLLM-initialized TNLS and TnALE across different datasets. 

\begin{table}[!h]
  \centering
  \scriptsize
  \caption{Performance comparison across different datasets for vanilla and tnLLM-initialized TNLS and TnALE. The values on the left give the lowest training objective function values and their corresponding testing objective function values. The values in [square brackets] give the number of evaluations required to first achieve the best training objective function value. For both metrics, a lower value is better. The least number of evaluations are highlighted in bold. \\}
  
  \label{tab:comparison_sec_2}
  \begin{tabular}{c c c c c c c c c c}
    \toprule[1.2pt]
    \multirow{2}{*}{\textbf{Data Type}} 
      & \multirow{2}{*}{} 
      & \multicolumn{4}{c}{\textbf{TNLS}} 
      & \multicolumn{4}{c}{\textbf{TnALE}} \\
    & 
      & \multicolumn{2}{c}{Vanilla} 
      & \multicolumn{2}{c}{tnLLM-init} 
      & \multicolumn{2}{c}{Vanilla} 
      & \multicolumn{2}{c}{tnLLM-init} \\
    \midrule[0.8pt]

    \multirow{2}{*}{\textbf{Images}} 
      & Train 
        & -0.66 
        & \multirow{2}{*}{[114]} 
        & -0.64 
        & \multirow{2}{*}{[\textbf{20}]} 
        & -0.65 
        & \multirow{2}{*}{[81]} 
        & -0.64 
        & \multirow{2}{*}{[\textbf{19}]} \\
      & Test  
        & -0.47 & 
        & -0.46 & 
        & -0.46 & 
        & -0.46 & 
        \\

    \midrule

    \multirow{2}{*}{\textbf{Videos}} 
      & Train 
        & -1.64 
        & \multirow{2}{*}{[484]} 
        & -1.64 
        & \multirow{2}{*}{[\textbf{21}]} 
        & -1.66 
        & \multirow{2}{*}{[254]} 
        & -1.65 
        & \multirow{2}{*}{[\textbf{20}]} \\
      & Test  
        & -1.72 & 
        & -1.71 & 
        & -1.72 & 
        & -1.70 & 
        \\

    \midrule

    \multirow{2}{*}{\textbf{Time‐series}} 
      & Train 
        & -0.45 
        & \multirow{2}{*}{[218]} 
        & -0.47 
        & \multirow{2}{*}{[\textbf{14}]} 
        & -0.47 
        & \multirow{2}{*}{[177]} 
        & -0.47 
        & \multirow{2}{*}{[\textbf{20}]} \\
      & Test  
        & -0.43 & 
        & -0.44 & 
        & -0.47 & 
        & -0.44 & 
        \\

    \bottomrule[1.2pt]
  \end{tabular}
\end{table}

\end{document}